%% file: acl_latex.tex
\pdfoutput=1

\documentclass[11pt]{article}

\usepackage[]{acl}

\usepackage{times}
\usepackage{latexsym}
\usepackage{booktabs}
\usepackage{comment}

\usepackage[T1]{fontenc}

\usepackage[utf8]{inputenc}

\usepackage{microtype}

\usepackage{inconsolata}
\usepackage{graphicx}               %
\usepackage{amsmath,amssymb}

\title{ARES: An Automated Evaluation Framework for Retrieval-Augmented Generation Systems}

\author{Jon Saad-Falcon \\
  Stanford University \thanks{Project started during research internship at Databricks} \\
  \texttt{jonsaadfalcon@stanford.edu} \\\And
  Omar Khattab \\
  Stanford University \\
  \texttt{okhattab@stanford.edu} \AND 
  Christopher Potts \\
  Stanford University \\
  \texttt{cgpotts@stanford.edu} \\\And 
  Matei Zaharia \\
  Databricks and UC Berkeley\\
  \texttt{matei@databricks.com}}

\begin{document}
\maketitle

\input{abstract}

\input{introduction}

\input{related_work}

\input{methods}

\input{experiments}

\input{results}

\input{future_work_and_conclusion}

\input{limitations}

\bibliography{anthology,custom}

\input{appendix}

\end{document}

%% file: abstract.tex
\begin{abstract}

Evaluating retrieval-augmented generation (RAG) systems traditionally relies on hand annotations for input queries, passages to retrieve, and responses to generate.
We introduce ARES, an \textit{Automated RAG Evaluation System}, for evaluating RAG systems along the dimensions of context relevance, answer faithfulness, and answer relevance.
By creating its own synthetic training data, ARES finetunes lightweight LM judges to assess the quality of individual RAG components.
To mitigate potential prediction errors, ARES utilizes a small set of human-annotated datapoints for prediction-powered inference (PPI).
Across eight different knowledge-intensive tasks in KILT, SuperGLUE, and AIS, ARES accurately evaluates RAG systems while using only a few hundred human annotations during evaluation.
Furthermore, ARES judges remain effective across domain shifts, proving accurate even after changing the type of queries and/or documents used in the evaluated RAG systems. We make our code and datasets publicly available \href{https://github.com/stanford-futuredata/ARES}{on Github}.

\end{abstract}

%% file: introduction.tex
\section{Introduction}

Retrieval-augmented generation (RAG) has become a prominent approach for building user-facing NLP applications, such as systems for question answering (QA), fact-checking, and customer support~\cite{petroni-etal-2021-kilt, wang2019superglue}. Typically, a RAG system consists of a retriever and a downstream language model (LM). Given a user question, the retriever finds relevant passages from a corpus 
and the LM uses these passages to generate a response. This formulation admits a multitude of choices: 
what retrieval model to use, %
how to divide the documents into retrieval chunks, %
and how to prompt %
or finetune %
the LM to use the retrieved information, to name only a few of the simplest design decisions.

The best design for a RAG system is not necessarily universal across data domains, corpus sizes, and cost/latency budgets. To tune their own RAG systems, practitioners traditionally need hand annotations for test questions, passages to retrieve (to assess the retriever), and responses to generate, labeled specifically for their target domain. Alternatively, they may evaluate different approaches in production by collecting human preferences that compare the candidate systems. Unfortunately, both of these strategies demand high expertise and impose considerable annotation costs.

Model-based evaluation is
an inexpensive
strategy to test generative output quality \cite{zheng2023judging}. For instance, the open-source RAGAS framework ~\cite{ragas2023} prompts an LM for evaluating the \textit{relevance} of retrieved information and the \textit{faithfulness} and \textit{accuracy} of generated responses. Unfortunately, such strategies currently rely for evaluation on a fixed set of heuristically hand-written prompts, offering little adaptability to various evaluation contexts and no guarantees about quality. %

To evaluate RAG systems rapidly and accurately, we propose ARES, the \textbf{A}utomated \textbf{R}AG \textbf{E}valuation \textbf{S}ystem.
ARES is the first automated RAG evaluation system to generate tailored LLM judges for each component of a RAG pipeline, leading to substantial boosts in evaluation precision and accuracy compared to existing approaches like RAGAS.
Furthermore, unlike existing RAG evaluation systems, ARES provides 
confidence intervals for its scoring
by leveraging prediction-powered inference (PPI; \citealt{angelopoulos2023predictionpowered}).
Given a corpus of documents and a RAG system, ARES reports three evaluation scores: context relevance (is the retrieved information pertinent to the test question), answer faithfulness (is the response generated by the language model properly grounded in the retrieved context), and answer relevance (is the response also relevant to the question). A good RAG system finds relevant contexts and generates answers that are both faithful and relevant.

Many existing RAG evaluation frameworks require substantial human annotations for scoring.
ARES significantly improves data efficiency during evaluation by only requiring three inputs: an in-domain passage set, a human preference validation set of approximately 150 annotated datapoints or more, and few-shot examples of in-domain queries and answers (e.g. five examples or more), which are used for prompting LLMs in synthetic data generation.

Given the corpus of in-domain passages, ARES proceeds in three stages. First, it leverages an LM to construct a synthetic dataset of question--answer pairs, derived from the passages in the corpus. 
Second, it defines three separate judge models to perform three classification tasks (context relevance, answer faithfulness, and answer relevance). These judges are lightweight models fine-tuned against a contrastive learning objective. 
Third, ARES scores the different RAG systems being assessed using prediction-powered inference (PPI; \citealt{angelopoulos2023predictionpowered}) to improve model-based evaluation accuracy and provide statistical confidence intervals for RAG scoring.
PPI utilizes a small set of human annotated datapoints for computing its confidence intervals; we designate this annotated set as our \textit{human preference validation set}, which is composed of approximately 150 annotated datapoints or more that designate both positive and negative examples for context relevance, answer faithfulness, and answer relevance.

We conduct extensive empirical evaluations, demonstrating that ARES accurately scores RAG systems across the six knowledge-intensive datasets in KILT and SuperGLUE, beating existing automated evaluation approaches like RAGAS by 59.3 and 14.4 percentage points on average across context relevance and answer relevance evaluation accuracy, respectively. 
Additionally, ARES accurately calculates answer hallucination occurrences in the AIS attribution dataset \cite{rashkin2022measuring}, predicting within 2.5 percentage points of the ground truth average for answer hallucinations.
Compared to annotation-based evaluation methods, ARES is substantially more accurate and efficient, requiring 78\% less annotations than the baseline approach.
We also find that ARES consistently distinguishes competitive RAG systems that are only a few points apart in ground-truth metrics. This precision enables ARES to guide the development and comparison of competitive approaches and configurations.

We make the ARES code and datasets publicly available \href{https://github.com/stanford-futuredata/ARES}{on Github}.

%% file: related_work.tex
\section{Related Work}

RAG~\citep{guu2020retrieval,lewis2020retrieval,khattab2021relevance,izacard2022few}) is now a common strategy for bolstering LLMs by combining them with retrieval systems. Through retrieval, RAG helps LM systems gather domain-specific knowledge, ground generations in factual information~\cite{shuster2021retrieval, huo2023retrieving}, and offer a degree of transparency or interpretability via citing sources~\cite{mialon2023augmented}.

Multiple LLM-based evaluation techniques have emerged for gauging LLM systems. This is essential for rapid deployment in new settings, where it is difficult to build a traditional benchmark dataset from scratch.
Early attempts at this use LLMs out of the box, as in MT-Bench and Chatbot Arena~\cite{zheng2023judging}.
AutoCalibrate~\cite{liu2023calibrating} seeks to align an LLM-judge with human preferences, leveraging a self-refinement prompt to iteratively improve the LLM judge. 
However, AutoCalibrate does not offer any statistical guarantees for the accuracy of its predictions.
Other work has used LLM prompting to evaluate system quality across natural language generation tasks, such as translation, summarization, and dialogue  \cite{kocmi2023large,fu2023gptscore, liu2303g, wang2023chatgpt}.

In the context of knowledge-intensive NLP tasks, LLMs have been explored for assessing attribution and factuality in LLMs \cite{min2023factscore, gekhman2023trueteacher, yue2023automatic}.
New guidelines like LongEval \cite{krishna-etal-2023-longeval} and datasets like Hagrid and ALCE \cite{kamalloo2023hagrid, gao2023enabling} provide resources for analyzing knowledge-intensive LLM pipelines.

The two most-closely related projects to ARES are EXAM~\cite{sander2021exam} and RAGAS~\cite{ragas2023}. To evaluate RAG systems, the EXAM metric estimates how many exam questions a reader (simulated as a QA system) can answer correctly based on the generated response. This requires a set of queries with several associated sub-questions each, which adds a burden that ARES does not bring. RAGAS is based on a handful of heuristic hand-written prompts. These offer little adaptability to new RAG evaluation settings (e.g., new corpora) and, as we show in our evaluation, substantially underperform ARES.

%% file: methods.tex
\input{Tables/Headline_Figure}

\section{ARES}
\label{sec:methods}

ARES proceeds in three stages (\autoref{fig:domain_adaptation_diagram}). 
There are three required inputs: an in-domain passage set, a human preference validation set of approximately 150 annotated datapoints (or more), and few-shot examples of in-domain queries and answers (five or more examples), which are used for prompting LLMs in synthetic data generation.
With our inputs prepared, we begin by generating synthetic queries (and their answers) from the passages in the target corpus. We then use these query--passage--answer triples to train LLM judges.
Subsequently, we apply these judges to any RAG system, scoring a sample of its in-domain query-document-answer triples, and use prediction-powered inference (PPI) with our human preference validation set to estimate a confidence interval for the quality of each RAG system.

\subsection{LLM Generation of Synthetic Dataset}

We generate synthetic queries and answers from the corpus passages using generative LLMs.
The generated data represent both positive and negative examples of query--passage--answer triples (e.g., relevant/irrelevant passages and correct/incorrect answers). 
For generation, the LLM uses our input set of few-shot examples with in-domain passages mapped to in-domain queries and answers; the model then generates a synthetic question and answer from a given in-domain passage, allowing us to create both positive and negative training examples. 
We include example prompts for generating synthetic queries and answers in \ref{sec:generate_synthetic_data}.

For creating our synthetic data, we primarily use on FLAN-T5~XXL (discussed in  \autoref{sec:models}). 
ARES works well with this model (see  \autoref{sec:results_and_analysis}) but our system can ultimately use another high-quality model for generating synthetic queries and answers.
We then filter out low-quality queries by testing if a given query can retrieve its original passage as the top result using its retriever.
This filtering approach has been used in previous work to isolate high-quality synthetic queries \cite{dai2022promptagator, saad2023udapdr}.

To generate negatives for fine-tuning our LLM judges, we rely on two novel strategies, generating the same number of negatives with each strategy:

\begin{enumerate}\setlength{\itemsep}{0pt}
    \item \textbf{Weak Negative Generation}: For context relevance negatives, we randomly sample in-domain passages unrelated to a given synthetic query. 
    For answer faithfulness and answer relevance negatives, we randomly sample synthetically-generated answers from other passages, which were created using FLAN-T5~XXL.
    \item \textbf{Strong Negative Generation}: For context relevance negatives, we randomly sample in-domain passages from the same document as the gold passage.
    For datasets in which multiple passages are not available for the same document, we use BM25 to retrieve the top-10 passages similar to the passage and sample from them for our context relevance strong negatives.  
    For answer faithfulness and answer relevance negatives, we prompt FLAN-T5 XXL (\autoref{sec:models}) to generate a contradictory answer using the few-shot prompt in \autoref{sec:flan_generate_synth_data}.
\end{enumerate}

In total, the number of negatives generated equals the number of positives generated for evaluating context relevance and answer relevance. %

\subsection{Preparing LLM Judges}

To prepare our RAG evaluation judges, we use our synthetic dataset to fine-tune DeBERTa-v3-Large judges (discussed in  \autoref{sec:models}) %
to evaluate %
three different capabilities
\cite{chen2023benchmarking, ragas2023}:

\begin{enumerate}\setlength{\itemsep}{0pt}
    \item \textbf{Context Relevance}: Is the passage returned relevant for answering the given query?
    \item \textbf{Answer Faithfulness}: Is the answer generated faithful to the retrieved passage, or does it contain hallucinated or extrapolated statements beyond the passage?
    \item \textbf{Answer Relevance}: Is the answer generated relevant given the query and retrieved passage?
\end{enumerate}

For each metric, a separate LLM with a binary classifier head is fine-tuned to classify positive and negative examples. 
For each concatenated query-document-answer, a single LLM judge must classify the triple as positive or negative for that judge's metric.
To fine-tune these judges, we use our human preference validation set to evaluate model improvement after each epoch, stopping when we have three epochs with no improvement in loss (see \autoref{sec:finetuning_configuration} for more information).

\subsection{Ranking RAG Systems with Confidence Intervals}
\label{sec:ranking_rag_with_ppi}

Once we have prepared our LLM judges, we need to use them to score and rank the competing RAG systems. 
To do this, ARES samples the in-domain query-document-answer triples produced by each RAG approach, and the judges label each triple, predicting their context relevance, answer faithfulness, and answer relevance.
By averaging the individual predicted labels for each in-domain triple, we calculate the RAG system performance across each of the three metrics.

In principle, we could simply report these average scores as quality metrics for each RAG system. However, these scores reflect entirely unlabeled data with predictions from a synthetically-trained LLM judge, and hence they may not be entirely accurate.
As an extreme alternative, we could use just the small human preference validation set discussed previously for evaluation, reporting the extent to which each RAG system agrees with (or deviates from) the human annotations.
However, an annotation-based evaluation approach would require labeling substantially more generative outputs from each RAG systems separately, which can be costly both in terms of time and financing.

To combine the benefits of both, and hence boost the precision of the evaluation, ARES uses \textit{prediction-powered inference} (PPI; \citealt{angelopoulos2023predictionpowered}) to predict the system scores. PPI is a recent statistical method that provides tighter confidence intervals on a small set of annotated datapoints (i.e., our validation set) by leveraging predictions on a much larger set of non-annotated datapoints.
PPI can leverage both the labeled datapoints and the ARES judge predictions on the non-annotated datapoints to construct confidence intervals for our RAG system's performance. %

To do this, PPI uses the LLM judges on the human preference validation set to learn a \textit{rectifier function} for constructing a confidence set of the ML model's performance, using each ML prediction in the larger non-annotated dataset. 
The confidence set can then be used to create a tighter confidence interval for the performance of the evaluated RAG system (e.g. its context relevance, answer faithfulness, or answer relevance accuracy individually) compared to simply using annotated outputs from the evaluated RAG system.
By bolstering the human preference validation set with the much larger set of datapoints with ML predictions, PPI can develop reliable confidence intervals for ML model performance that beat previous classical inference approaches.

The PPI rectifier function allows us to estimate the errors of the LLM judge and generate confidence bounds for the success and failure rates of the RAG system, estimating context relevance, answer faithfulness, and answer relevance performance.
Additionally, PPI allows us to estimate confidence intervals with a selected level of probability; for our experiments, we use a standard 95\% alpha (probability) for our confidence interval.

With the accuracy confidence interval for each component of the RAG, we find the midpoint of each confidence interval and use the midpoints to rank the RAG systems.
With our ranking, we can compare different RAG systems, as well as different configurations of the same RAG system, to find the best-performing approach for a given domain.

%% file: Tables/Headline_Figure.tex
\begin{figure*}[tp!]
   \centering
   \includegraphics[width=0.9\linewidth]{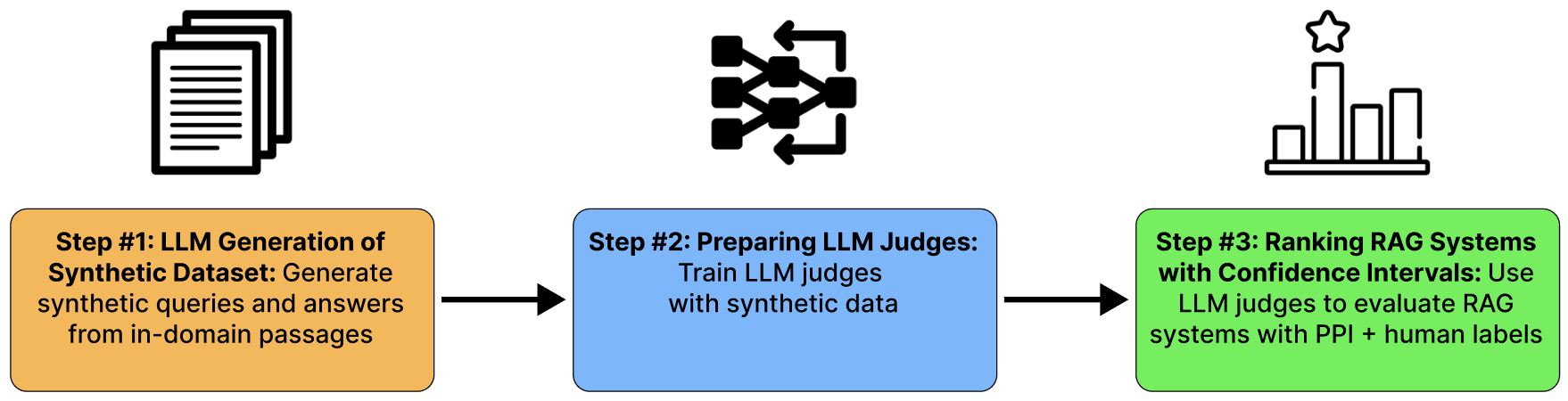}
   \caption{\textbf{Overview of ARES}:
   As inputs, the ARES pipeline requires an in-domain passage set, a human preference validation set of 150 annotated datapoints or more, and few-shot examples of in-domain queries and answers (five or more), which are used for prompting LLMs in synthetic data generation.
   To prepare our LLM judges for evaluation, we first generate synthetic queries and answers from the corpus passages. 
   Using our generated training triples and a constrastive learning framework, we fine-tune an LLM to classify query--passage--answer triples in three different criteria: context relevance, answer faithfulness, and answer relevance.
   Finally, we use the LLM judges to score RAG systems and generate confidence bounds for the ranking using PPI and the human preference validation set. 
   }
   \label{fig:domain_adaptation_diagram}
\end{figure*}

%% file: experiments.tex
\section{Experiments}

\subsection{Models}
\label{sec:models}

For our fine-tuned judges,
ARES relies on generating cheap but quality synthetic queries and answers using LLMs.
For generating our synthetic datasets, we use FLAN-T5 XXL \cite{chung2022scaling}.
We selected DeBERTa-v3-Large \cite{he2021debertav3} for our fine-tuned LLM judge.
Our fine-tuned LLM judges allow us to rank RAG systems without relying on external APIs, solely using few-shot prompts and deployable LLMs on commercial GPUs.

For our in-context learning baseline, we use OpenAI's \textit{gpt-3.5-turbo-16k}, version 10/23, \cite{brown2020language} in a zero/few-shot setting.
For similarity search over in-domain passages, we use FAISS IndexFlatL2 for indexing \cite{johnson2019billion} and OpenAI's \textit{text-embedding-ada-002} for generating embeddings.
We use simlarity search over in-domain passages to filter our synthetic queries that cannot retrieve the passage from which they were generated.
We use version 0.0.18 of RAGAS in our experiments \cite{ragas2023}.

\subsection{Datasets}
\label{sec:datasets}

Our core experimental goal is to provide a rich picture of where ARES can be applied effectively.
To test across multiple types of queries, documents, and answers, 
we selected all the datasets from the widely-used KILT and SuperGLUE benchmarks for which RAG is appropriate.

From KILT \cite{petroni-etal-2021-kilt}, we use Natural Questions (NQ), HotpotQA, FEVER, and Wizards of Wikipedia (WoW) \cite{kwiatkowski2019natural, yang2018hotpotqa, fever-2023-fact, dinan2018wizard}. 
Each dataset uses Wikipedia passages but the queries and answers offer a range of applications.
Both NQ and HotpotQA feature direct questions and expect short answers, but NQ uses single passages for reasoning while HotpotQA requires multiple passages for reasoning.
Furthermore, FEVER focuses on fact-verification, determining if a passage supports or refutes a given statement, and expects an output of ``SUPPORTS'' or ``REFUTES''.
WoW seeks to evaluate dialogue agents by mapping user dialogue to relevant Wikipedia passages before a chatbot generates a paragraph-length chat response incorporating passage knowledge.

From SuperGLUE \cite{wang2019superglue}, we use  MultiRC and ReCoRD \cite{khashabi2018looking, zhang2018record}.
MultiRC focuses on direct questions for seven different domains (News, Wikipedia articles, articles on society/law/justice, articles on history/anthropology, elementary school science textbooks, 9/11 reports, and fiction).
ReCoRD focuses on determining the placeholder entity in a statement, focusing on news articles from CNN and the Daily Mail.
For MultiRC and ReCoRD, we create open-domain versions of their tasks. 
For MultiRC, we perform retrieval over its seven sets of domain passages. For ReCoRD, we perform retrieval over its news article passages.

The efficacy of ARES relies on its ability to rank different RAG systems while only using a human preference validation set and domain-targeted LLM judges.
To test the limits of ARES, we need to simulate the existence of many RAG systems that are separated by small accuracy margins on our evaluation metrics.
For this, we create systems using artificial query-passage-answer triples, in which we empirically know the positive and negative examples of the mock RAG system.
We generate these mock splits of the given datasets by selecting
(1) The positive and negative query-passage matches for context relevance, and
(2) the positive and negative query-passage-answer matches for answer relevance.
We include positive and negative examples from our evaluation sets in \autoref{tab:positive_negative_examples}.

For our positive triples, we can simply use the KILT and SuperGLUE examples without any alteration.
For gathering negative query-passage pairs and query-passage-answer triples, we randomly sample passages and answers from either: the same Wikipedia document or an entirely random Wikipedia document.
This sampling allows us to artificially create mock RAG systems for testing ARES.
By sampling both related and unrelated documents/answers, we hope to better gauge the efficacy of ARES in judging RAG outputs.

We do not evaluate answer faithfulness for KILT and SuperGLUE datasets since we do not have human-annotated hallucinated answers to use for evaluation. 
However, we do test the ARES framework on real attribution datasets in Section \ref{sec:ais}. 

Using the validation subsets for each KILT and SuperGLUE dataset, we create nine different dataset splits, ranging from 70\% success rate to 90\% success rate for each of the evaluated RAG criteria; each dataset is separated by 2.5\% accuracy points (e.g. 70.0\%, 72.5\%, 75.0\%, \ldots, 90.0\%).
Each split also represents a different mock RAG system.
Since we know the success percentages of each dataset split, we know the appropriate ranking of each mock RAG system.
This allows us to test ARES success at both scoring and ranking the mock RAG systems appropriately across the three evaluation criteria.

\subsection{Metrics}

To calculate the correlation between the correct ranking and the ARES ranking, we use the Kendall rank correlation coefficient or Kendall's $\tau$:
\begin{equation*} 
\small
{\tau = \frac{(\# \, \text{of concordant pairs}) \\ - (\# \, \text{of discordant pairs})}{\# \, \text{ of pairs total}}}
\end{equation*}
Concordant pairs are defined as two ordinal values in the ranking where the earlier value in the sequence is lower than the later value in the sequence. Discordant pairs are defined as two ordinal values in the ranking where the earlier value in the sequence is greater than or equal to the later value in the sequence.
A Kendall's $\tau$ greater than 0.9 is considered successful but it ranges from 0.0 to 1.0.

In development, researchers and engineers will be comparing different RAG configurations through individual pairwise comparisons of model choices, retriever selection, and document preprocessing.
We want to make sure that ARES has satisfactory accuracy in pairwise comparisons across a variety of performance gaps between RAG systems. 
Kendall's $\tau$ is explicitly designed for measuring the accuracy of such pairwise comparisons, calculating the correlation between a perfectly accurate pairwise ranking and an experimental pairwise ranking.
Thus, it is a popular and widespread metric used in information retrieval, allowing developers to evaluate ranking systems empirically.
Therefore, we believe Kendall's tau and  prediction accuracy provide meaningful metrics for testing the efficacy of ARES as a RAG evaluation system.

%% file: results.tex
\section{Results \& Analysis}
\label{sec:results_and_analysis}

\subsection{ARES Ranking}

\input{Tables/ARES_Ranking_vs_GPT3.5}

\autoref{tab:ares_vs_gpt3.5} summarizes our main evaluation of ARES (with DeBERTa-v3-Large as the pretrained basis for the judges). We compare against RAGAS (version 0.0.18) and a baseline few-shot prompted GPT-3.5 judge (\textit{gpt-3.5-turbo-16k}).
For the few-shot GPT-3.5 judge, we provide few-shot examples for guiding predictions; the prompts are included in Appendices \ref{sec:gpt_prompting_for_context_relevance_scoring}, \ref{sec:gpt_prompting_for_answer_faithfulness_scoring}, and \ref{sec:gpt_prompting_for_answer_relevance_scoring}.
For both ARES and the GPT-3.5 judge baseline, we augment the LLM with PPI, using a 300-datapoint human preference validation set to rectify the ML predictions and produce confidence intervals.

Across almost all settings across the datasets from KILT and SuperGLUE,
ARES provides a more accurate ranking of RAG systems than RAGAS. ARES averages a Kendall's $\tau$ \textit{0.065 higher for context relevance} and \textit{0.132 higher for answer relevance than RAGAS}.
Additionally, the LLM-judge is substantially more accurate than RAGAS at predicting context relevance and answer relevance of a query-passage-answer triple.
For context relevance, ARES with a fine-tuned LLM-judge is \textit{59.9 percentage points higher than RAGAS} while for answer relevance, our system is \textit{14.4 percentage points higher than RAGAS}. 
Overall, ARES provides a more accurate system for automatically evaluating RAG configurations than RAGAS by leveraging domain-adaptive techniques for prompting and training as well as utilizing PPI to bolster model predictions.

As an additional comparison, we also include the Kendall's $\tau$ for RAG ranking with the ARES LLM judge without PPI; for all datasets tested, PPI improved the ranking prediction accuracy of the fine-tuned LLM judge.
Furthermore, we included a sampled annotations configuration, in which we sampled 150-datapoints from each mock RAG system, totalling 1,350 annotations.
Even with all these annotations, the Kendall's $\tau$ for ARES is 0.08 higher on average, across both context and answer relevance, compared to sampled annotations, despite using 78\% less annotations.
In sum, ARES proves significantly more data-efficient with human annotations while being more accurate at scoring than standard sampled annotation methods.

Compared to the GPT-3.5 judge, ARES provides a more accurate ranking of the RAG systems than the GPT-3.5 judge, averaging a Kendall's tau 0.06 higher over both context relevance and answer relevance.
Between the judge configurations,
the fine-tuned LLM judge of ARES can more precisely distinguish between RAG systems and guide configuration decisions surrounding document splitting, retriever selection, and generative LLM choice.
However, while the fine-tuned LLM judge had a higher Kendall's tau on average, the GPT-3.5 judge is more readily deployable and does not require any additional fine-tuning.
The GPT-3.5 judge does come with its own querying costs, which can vary based on the date of querying as well as the total tokens used in evaluation.

We also wanted to better understand the importance of human annotations for ARES. To this end, we conducted two sets of experiments. First, we used ARES with human annotation sets ranging in size from 25 to 400 and found that 150 is the minimum number required (\autoref{tab:ppi_count}). Second, we explored whether GPT-4 generations could replace human annotations entirely, finding that GPT-4 is less good than humans in this role, though the idea arguably has promise (\autoref{tab:gpt4_labels}).

\subsection{ARES Performance on AIS}
\label{sec:ais}

\begin{table}[htp!]
\centering
\small
\begin{tabular}{lcc}
\toprule
                                                                               & \multicolumn{1}{l}{\textbf{WoW}} & \multicolumn{1}{l}{\textbf{CNN / DM}} \\
                                                                               \midrule
ARES Split Prediction                                                                & 0.478                           & 0.835                                \\
Correct Positive/Negative Split                                                                  & 0.458                           & 0.859                                \\
ARES Judge Accuracy                 & 62.5\%                           & 84.0\%                                \\
Evaluation Set Size                & 707                              & 510                                   \\
Human Preference Data Size & 200 & 200  \\       
\bottomrule
\end{tabular}
\caption{ARES Results on the AIS benchmark}
\label{tab:ais_results}
\end{table}

To evaluate whether ARES can effectively gauge answer faithfulness in real RAG systems, we tested ARES on the AIS attribution benchmark \cite{rashkin2022measuring}. 
In AIS, we selected the Wizards of Wikipedia (WoW) and CNN/DM datasets; the other benchmark datasets involve either table reasoning (ToTTo) or focus on passage summarization (QRECC) so we excluded them.
In WoW and CNN/DM, each evaluation example includes a query, a retrieved passage, and a generated answer (which is either faithful or non-attributed to the retrieved passage).

\autoref{tab:ais_results} summarizes our AIS results. We found that ARES can effectively score the AIS datasets, getting within 2.5 accuracy points of the correct scores. Furthermore, for scoring each system, we only use 200 annotated datapoints for our human preference validation set.
Our results on AIS demonstrate the ability of ARES to reliably distinguish faithful and hallucinated answers in real-world RAG systems.

\subsection{ARES Ranking of Existing RAG Systems}

We also wanted to evaluate whether ARES can score and rank existing RAG systems across both context relevance and answer relevance. 
For evaluation, we selected the NQ, WoW, and FEVER datasets from KILT.
We consider the answer generations to be correct if they contained the KILT answer in their output.
For our RAG systems, we selected three different retrievers (BM25, OpenAI Ada embeddings with cosine similarity search, and ColBERTv2 \cite{santhanam-etal-2022-colbertv2}) and three different generative LLMs (MPT-7b-Instruct \cite{MosaicML2023Introducing}, GPT-3.5-Turbo, and GPT-4).
Additionally, we include the Facebook RAG model \cite{lewis2020retrieval}, which uses a DPR retriever \cite{karpukhin2020dense} and BART sequence-to-sequence model \cite{lewis2019bart}. 
During retrieval, each RAG system only retrieves one passage to assist generation.

In \autoref{tab:ares_real_systems}, we found that ARES can reliably score and rank RAG systems in real-world applications, averaging a Kendall's tau of 0.91 for context relevance and 0.97 for answer relevance.
Compared to RAGAS, ARES is 0.16 higher for context relevance and 0.15 higher for answer relevance, on average.
ARES also provided accurate confidence bounds for its predictions, capturing the ground truth average outcomes for context relevance and answer relevance more than 95\% of the time; on average, the PPI confidence intervals were 7.4 points wide for context relevance and 6.1 points wide for answer relevance (see \autoref{fig:nq_context_relevance_real_rag} and \autoref{fig:nq_answer_relevance_real_rag} for ARES vs. RAGAS).
Among the models tested, the best performing retriever was ColBERTv2 while the best performing generative LLM was GPT-4.

\subsection{Strengths and Limits of Cross-Domain Applications}

The generalizability of the LLM judge used in ARES is critical for deploying our framework in specialized domains, particularly domains where in-domain queries, documents, and answers are difficult to gather.
Therefore, we wanted to test how the LLM judges used in ARES would be affected by three domain shifts: change in \textit{query type} from training to test (e.g. NQ to FEVER), change in \textit{document type} from training to test (e.g. NQ to MultiRC), and change in both \textit{query and document type} (e.g. NQ to ReCoRD).

In \autoref{tab:cross_domain}, we found that the fine-tuned LLM judges used in ARES proved successful in cross-domain applications. 
Across all settings, we found that LLM judges in ARES had strong generalizability, even when only using 300 datapoints in our human preference validation set for PPI.
Furthermore, we found that even when the LLM judge's accuracy suffered in cross-domain applications, PPI helped mitigate the loss in accuracy and still allow ARES to be successful.
Additional examples for PPI also continued to boost cross-domain ARES performance in subsequent tests.

While LLM judges in ARES were successful in cross-domain applications for KILT and SuperGLUE, LLM judges are unable to generalize when making more drastic shifts in domain, such as: switching languages (e.g. English to Spanish, German, and other languages), switching from text to code (e.g. questions + passages to coding functions + documentation), and switching from retrieving text to extraction of entities, webpages, or citations.

To test cross-lingual transfer, we used the XGLUE datasets \cite{Liang2020XGLUEAN}; a LLM judge fine-tuned on NQ achieved a Kendall's tau of 0.33 over both context relevance and answer relevance scoring for XGLUE.
To test text-to-code, we used CodeSearchNet \cite{husain2019codesearchnet}; an LLM judge fine-tuned on NQ achieved a Kendall's tau of 0.28 over both context relevance and answer relevance scoring for CodeSearchNet.
To test extraction task generalizability, we used T-Rex from KILT \cite{elsahar2018t, petroni-etal-2021-kilt}; an LLM judge fine-tuned on NQ achieved a Kendall's tau of 0.38 over both context relevance and answer relevance scoring for T-Rex.
Each cross-domain shift requires in-domain passages and few-shot query examples for reconfiguring ARES judges.

%% file: Tables/ARES_Ranking_vs_GPT3.5.tex
\begin{table*}[ht]
\small
\centering
\setlength{\tabcolsep}{1.5pt}
\begin{tabular}{lrrrrrrrrrrrr}
\toprule
 & \multicolumn{12}{c}{\textbf{ARES Ranking of Pseudo RAG Systems}}\\
 \cmidrule(l{7pt}r{7pt}){2-13}
 & \multicolumn{2}{c}{NQ}  & \multicolumn{2}{c}{HotpotQA} & \multicolumn{2}{c}{WoW} & \multicolumn{2}{c}{FEVER} & \multicolumn{2}{c}{MultiRC}  & \multicolumn{2}{c}{ReCoRD}\\
 \cmidrule(l{7pt}r{7pt}){2-13}
 & \multicolumn{1}{c}{C.R}& \multicolumn{1}{c}{A.R.}  & \multicolumn{1}{c}{C.R}& \multicolumn{1}{c}{A.R.}  & \multicolumn{1}{c}{C.R}& \multicolumn{1}{c}{A.R.}  & \multicolumn{1}{c}{C.R}& \multicolumn{1}{c}{A.R.}  & \multicolumn{1}{c}{C.R}& \multicolumn{1}{c}{A.R.}  & \multicolumn{1}{c}{C.R}& \multicolumn{1}{c}{A.R.} \\ \midrule
\begin{tabular}[c]{@{}l@{}}Kendall's Tau for \\ Sampled Annotations \end{tabular} & 0.83 & 0.89 & 0.78 & 0.78 & 0.78 & 0.83 & \textbf{0.89} & \textbf{0.89} & 0.83 & 0.83 & 0.72 & 0.94  \\ \midrule
\begin{tabular}[c]{@{}l@{}}Kendall's Tau \\ for RAGAS \end{tabular} &  0.89 &0.89&\textbf{0.94}  &0.89&  0.94 & 0.94&  0.72 &0.61&0.83&  \textbf{0.94} &\textbf{0.89}&0.44  \\ \midrule
\begin{tabular}[c]{@{}l@{}}Kendall's Tau \\ for GPT-3.5 Judge\end{tabular} & 0.89& 0.94& 0.67& \textbf{0.94}  & 0.94& 0.89& 0.78& {0.78}& 0.83& {0.89}& {0.83}  & \textbf{0.94}\\
\midrule
\begin{tabular}[c]{@{}l@{}}Kendall's Tau for \\ ARES LLM Judge \end{tabular} & 0.89  & \textbf{1.0} & 0.89 & \textbf{0.94} & 0.94 & \textbf{1.0} & 0.83 & 0.72 & \textbf{0.94} & 0.83 & 0.78 & 0.83 \\ \midrule
\begin{tabular}[c]{@{}l@{}}Kendall's Tau \\ for ARES \end{tabular} & \textbf{0.94}  & \textbf{1.0}& \textbf{0.94}  & \textbf{0.94}  & \textbf{1.0}& \textbf{1.0}& \textbf{0.89}  & {0.78}  & \textbf{0.94}  & {0.89}  & {0.83}  & {0.89}  \\ \midrule
RAGAS Accuracy &  31.4\% &  71.2\% & 17.2\% & 76.0\%  &36.4\%&  77.8\% &23.7\%& 69.2\%&  16.1\% &  75.0\% &  15.0\% &  72.8\%\\ \midrule
GPT-3.5 Judge Accuracy & 73.8\%& 95.5\%& 75.3\%& 71.6\%& 84.3\%& 85.2\%& 60.4\%& 59.6\%& 72.4\%& 60.3\%& 81.0\%& 65.8\% \\ \midrule
ARES Accuracy & 79.3\%& 97.2\%& 92.3\%& 81.3\%& 85.7\%& 96.1\%& 88.4\%& 78.5\%& 85.8\%& 82.7\%& 67.8\%& 92.3\% \\ \bottomrule 
\end{tabular}
\caption{\textbf{ARES Ranking with Fine-tuned LLM Judges vs. Sampled Annotations, RAGAS and GPT-3.5 Judge}: 
For scoring context relevance and answer relevance (C.R. and A.R. in the table, respectively), we compare ARES with our fine-tuned LLM judges against sampled annotations benchmark, RAGAS, %
and a few-shot GPT-3.5 judge.
For our sampled annotations, we gather 150 annotated datapoints from each mock RAG system and use those labels to score the system.
RAGAS also uses GPT-3.5 as its judge but it uses few-shot prompts that are not targeted for each evaluation domain.
Overall, we found that ARES ranked RAG systems more accurately than RAGAS and GPT-3.5 across all the explored datasets.
The Kendall's tau for ARES was \textit{0.065 higher on average for scoring context relevance} and \textit{0.132 higher on average for scoring answer relevance} than RAGAS.
Additionally, we include the Kendall's taus for the ARES LLM Judge without PPI and found that PPI further boosted the ranking accuracy of the judge across the board.
We selected GPT-3.5 instead of GPT-4 due to the lower financial costs required to run. %
For PPI in both ARES and the GPT-3.5 judge, we used 300 human annotations for our human preference validation set.
The prompts used for the GPT-3.5 judges are included in Sections \ref{sec:gpt_prompting_for_context_relevance_scoring}, \ref{sec:gpt_prompting_for_answer_faithfulness_scoring}, and \ref{sec:gpt_prompting_for_answer_relevance_scoring}.}
\label{tab:ares_vs_gpt3.5}
\end{table*}

%% file: future_work_and_conclusion.tex
\section{Conclusion}

In this work, we present ARES, a novel automated evaluation framework for retrieval-augmented generation (RAG).
ARES offers a novel training pipeline for fine-tuning lightweight LLM judges on synthetically generated queries and answers.
ARES can evaluate each component of a RAG system separately to help improve system understanding and create targeted solutions, and it requires only minimal human annotations.
For the eight different datasets in KILT, SuperGLUE, and AIS requiring RAG-based solutions, we found that ARES can accurately score and rank RAG systems based on context relevance, answer faithfulness, and answer relevance scores, beating the existing RAGAS automated evaluation framework. 

ARES is a flexible framework, and there may be variants of it that are even more powerful than the ones we explored here. Avenues to explore include GPT-4 as a replacement for human labeling (\autoref{tab:gpt4_labels}), more robust techniques for the synthetic datasets used in fine-tuning LLM judges, 
utilizing logits in LLM judge prediction to improve PPI confidence intervals, and testing more sophisticated LLMs as fine-tuned judges for ARES.

%% file: limitations.tex
\section{Limitations}

ARES relies on a small set of annotations in the human preference validation set (roughly 150-300 datapoints but more is better). 
These annotations often require an annotator familiar with the RAG system's domain application.
While these annotations can be easy to generate for general-domain applications, more specialized domains, such as law, medicine, and finance, may require annotators with specialized expertise.

The LLMs used in ARES benefit substantially from GPU-based hardware with substantial storage. 
In ARES, DeBERTa-v3-Large (304M) and FLAN-T5-XXL (11.3B) required GPUs with about 32GB of memory to run, taking several hours for fine-tuning and generation, respectively.
While commercial GPUs are widely available, they are not easily accessible to all NLP researchers and practitioners due to their costs.

Additionally, all of the datasets used in our evaluation of ARES are in English, a well-resourced language with abundant annotations.
Future work should explore how ARES can be employed in other languages by utilizing different LLMs for the ARES judge and the synthetic data generation.
This can help us better understand the strengths and weaknesses of the current ARES framework.

%% file: appendix.tex
\appendix

\section{Appendix}
\label{sec:appendix}

\subsection{Fine-tuning Configuration for LLM Judges}
\label{sec:finetuning_configuration}

For our loss function used in LLM judge training, we selected cross-entropy loss using Adam \cite{kingma2017adam}. 
For our classification head, we use a single linear classification layer and apply a 0.1 dropout to the input, which is the final hidden state of the [CLS] token.
For our learning schedule, we use linear warmup and linear decay \cite{howard-ruder-2018-universal} with a 5e-6 learning rate and a 32 training batch size across all experimental configurations.

\subsection{GPT Prompting for Context Relevance Scoring}
\label{sec:gpt_prompting_for_context_relevance_scoring}

For the NQ, HotpotQA, MultiRC, and ReCoRD datasets, we use 8 few-shot examples with the following prompt to score context relevance:

\begin{itemize}
    \item Given the following question and document, you must analyze the provided document and determine whether it is sufficient for answering the question. 
    In your evaluation, you should consider the content of the document and how it relates to the provided question. 
    Output your final verdict by strictly following this format: "[[Yes]]" if the document is sufficient and "[[No]]" if the document provided is not sufficient. 
    Do not provide any additional explanation for your decision.
    
    Question: <\textit{few-shot example here}>
    
    Document: <\textit{few-shot example here}>
    
\end{itemize}

For FEVER, we use the following prompt to score context relevance:

\begin{itemize}
    \item You are an expert fact-checking agent. Given the following statement and document, you must analyze the provided document and determine whether it is sufficient for determining the statement's factuality. In your evaluation, you should consider the content of the document and how it relates to the provided statement's factuality. Output your final verdict by strictly following this format: "[[Yes]]" if the document is sufficient and "[[No]]" if the document is not sufficient. Do not provide any additional explanation for your decision.
    
    Statement: <\textit{few-shot example here}>
    
    Document: <\textit{few-shot example here}>
    
\end{itemize}

For WoW, we use the following prompt to score context relevance:

\begin{itemize}
    \item You are an expert dialogue agent. Given the following dialogue and document, you must analyze the provided document and determine whether it is relevant for responding to the dialogue. In your evaluation, you should consider the content of the document and how it relates to the provided dialogue. Output your final verdict by strictly following this format: "[[Yes]]" if the document is relevant and "[[No]]" if the document provided is not relevant. Do not provide any additional explanation for your decision.

    Dialogue: <\textit{few-shot example here}>
    
    Document: <\textit{few-shot example here}>
    
\end{itemize}

\subsection{GPT Prompting for Answer Faithfulness Scoring}
\label{sec:gpt_prompting_for_answer_faithfulness_scoring}

For the NQ, HotpotQA, MultiRC, and ReCoRD datasets, we use 8 few-shot examples with the following prompt to score answer faithfulness:

\begin{itemize}
    \item Given the following question, document, and answer, you must analyze the provided answer and determine whether it is faithful to the contents of the document. The answer must not offer new information beyond the context provided in the document. The answer also must not contradict information provided in the document. Output your final verdict by strictly following this format: "[[Yes]]" if the answer is faithful to the document and "[[No]]" if the answer is not faithful to the document. Do not provide any additional explanation for your decision.

    Question: <\textit{few-shot example here}>
    
    Document: <\textit{few-shot example here}>

    Answer: <\textit{few-shot example here}>

\end{itemize}

For FEVER, we change the word "question" in the prompt to "statement". 
For WoW, we change the word "question" in the prompt to "dialogue".

\subsection{GPT Prompting for Answer Relevance Scoring}
\label{sec:gpt_prompting_for_answer_relevance_scoring}

For the NQ, HotpotQA, MultiRC, and ReCoRD datasets, we use 8 few-shot examples with the following prompt to score answer relevance:

\begin{itemize}
    \item Given the following question, document, and answer, you must analyze the provided answer and document before determining whether the answer is relevant for the provided question. In your evaluation, you should consider whether the answer addresses all aspects of the question and provides only correct information from the document for answering the question. Output your final verdict by strictly following this format: "[[Yes]]" if the answer is relevant for the given question and "[[No]]" if the answer is not relevant for the given question. Do not provide any additional explanation for your decision.

    Question: <\textit{few-shot example here}>
    
    Document: <\textit{few-shot example here}>

    Answer: <\textit{few-shot example here}>
\end{itemize}

For FEVER, we change the word "question" in the prompt to "statement". 
For WoW, we change the word "question" in the prompt to "dialogue".

\subsection{Prompting for Generation of Synthetic Queries and Answers}
\label{sec:flan_generate_synth_data}

To generate synthetic queries and answers using FLAN-T5, we use the following prompt and provide 5 few-shot examples:

\begin{itemize}
    \item Example N

    Question: <\textit{few-shot example here}>

    Document: <\textit{few-shot example here}>

    Answer: <\textit{few-shot example here}>
    
\end{itemize}

We use the same prompting structure for generating incorrect or contradictory answers; we simply swap out the few-shot examples to be incorrect or contradictory instead.

\subsection{Synthetic Query and Answer Generation}
\label{sec:generate_synthetic_data}

For generating our synthetic questions, we use the following prompt for FLAN-T5 XXL:

\begin{itemize}
    \item Example \#1
    
    Document: <\textit{few-shot example here}>
    
    Query: <\textit{few-shot example here}>
    
    Example \#2
    
    Document: <\textit{few-shot example here}>
    
    Query: <\textit{few-shot example here}>

    Example \#3
    
    Document: <\textit{few-shot example here}>
    
    Query: <\textit{few-shot example here}>

    Example \#4
    
    Document: <\textit{in-domain passage}>
    
    Query: 
\end{itemize}

For generating our synthetic answers, we use the following prompt for FLAN-T5 XXL:

\begin{itemize}
    \item Example \#1

    Query: <\textit{few-shot example here}>
    
    Document: <\textit{few-shot example here}>
    
    Answer: <\textit{few-shot example here}>
    
    Example \#2
    
    Query: <\textit{few-shot example here}>
    
    Document: <\textit{few-shot example here}>
    
    Answer: <\textit{few-shot example here}>

    Example \#3
    
    Query: <\textit{few-shot example here}>
    
    Document: <\textit{few-shot example here}>
    
    Answer: <\textit{few-shot example here}>

    Example \#4
    
    Query: <\textit{synthetic query here}>
    
    Document: <\textit{in-domain passage here}>
    
    Answer:
\end{itemize}

\input{Tables/NQ_ContextRelevance}

\input{Tables/NQ_AnswerRelevance}

\input{Tables/ppi_comparison_table}

\input{Tables/GPT4_Labeling}

\input{Tables/ARES_on_Real_RAG}

\input{Tables/Cross_Domain}

\begin{table*}[]
\tiny
\centering
\begin{tabular}{|l|l|l|l|l|}
\hline
\multicolumn{1}{|c|}{\textbf{Query}}                                                                                               & \multicolumn{1}{c|}{\textbf{Passage}}                                                                                                                                                                                                                                                                                                                                                                                                                                                                                                                                                                                                                                                                                                                                                                                                                                                                                                                                                                                    & \multicolumn{1}{c|}{\textbf{Answer}}                                                                        & \multicolumn{1}{c|}{\textbf{\begin{tabular}[c]{@{}c@{}}Context\\ Relevance\end{tabular}}} & \multicolumn{1}{c|}{\textbf{\begin{tabular}[c]{@{}c@{}}Answer\\ Relevance\end{tabular}}} \\ \hline
\begin{tabular}[c]{@{}l@{}}How can a ball that is not \\ moving possess energy \\ of position?\end{tabular}                        & \begin{tabular}[c]{@{}l@{}}Mechanical energy is a combination of the energy of motion or position. \\ This type of energy describes objects that are moving or could move. \\ A moving ball can have energy from motion. An arrow can also have \\ the energy of motion. Both are types of mechanical energy.\end{tabular}                                                                                                                                                                                                                                                                                                                                                                                                                                                                                                                                                                                                                                                                                               & \begin{tabular}[c]{@{}l@{}}The ball holds \\ mechanical energy\end{tabular}                                 & 1                                                                                         & 1                                                                                        \\ \hline
\begin{tabular}[c]{@{}l@{}}Who has a Jimmy\\  Stewart-like quality \\ of quiet trust?\end{tabular}                                 & \begin{tabular}[c]{@{}l@{}}One look at Fred Rooney, and you just know he's the good guy. \\ A trace of childish innocence in his face gives the lanky \\ Bethlehem lawyer a Jimmy Stewart-like quality of quiet trust. \\ In black jeans and button-down shirt, he's a kind of folk hero \\ in the south Bethlehem melting pot where he's crafted a law \\ practice catering to working-class families - mostly Latino - \\ in the shadow of the hulkish remnants of Bethlehem Steel.\end{tabular}                                                                                                                                                                                                                                                                                                                                                                                                                                                                                                                       & Fred Rooney                                                                                                 & 1                                                                                         & 1                                                                                        \\ \hline
\begin{tabular}[c]{@{}l@{}}Before he murder the \\ doctor and Ralph Smith, \\ where did the stepfather \\ reside?\end{tabular}     & \begin{tabular}[c]{@{}l@{}}Surviving being shot and stabbed at the end of the previous film , \\ the stepfather has been institutionalized in Puget Sound, Washington since , \\ spending his time building model houses in the workshop.\\ Assigned a new doctor named Joseph Danvers the stepfather \\ begins confiding in him to gain his trust , ultimately murdering \\ the doctor during a session by stabbing him in the neck with a \\ blade smuggled out of the workshop . After killing Danvers the stepfather \\ beats a suspicious guard named Ralph Smith to death with his own nightstick \\ with only two strikes and takes his uniform , successfully \\ sneaking out of the sanitarium . Checking into a hotel after robbing and \\ murdering a traveling salesman the stepfather alters his appearance , \\ takes the name Doctor Gene F. Clifford from the newspaper obituaries \\ and travels to Palm Meadows , Los Angeles after seeing an ad for it on \\ an episode of Dream House .\end{tabular} & Los Angeles                                                                                                 & 1                                                                                         & 0                                                                                        \\ \hline
\begin{tabular}[c]{@{}l@{}}What was the name of the \\ 2006 film about Pushkin's death, \\ and who portrayed Pushkin?\end{tabular} & \begin{tabular}[c]{@{}l@{}}After arriving in New York City, Einstein was taken to various places and \\ events, including Chinatown, a lunch with the editors of the New York \\ Times, and a performance of Carmen at the Metropolitan Opera, \\ where he was cheered by the audience on his arrival. \\ During the days following, he was given the keys to the city by Mayor \\ Jimmy Walker and met the president of Columbia University, who \\ described Einstein as "The ruling monarch of the mind." Harry \\ Emerson Fosdick, pastor at New York's Riverside Church, gave \\ Einstein a tour of the church and showed him a full-size statue that \\ the church made of Einstein, standing at the entrance.\end{tabular}                                                                                                                                                                                                                                                                                        & \begin{tabular}[c]{@{}l@{}}Vasily Szaitsev portrayed \\ Pushkin in the film \\ Pushkin Returns\end{tabular} & 0                                                                                         & 0                                                                                        \\ \hline
\end{tabular}
\caption{Positive and Negatives Evaluation Examples}
\label{tab:positive_negative_examples}
\end{table*}

%% file: Tables/NQ_ContextRelevance.tex
\begin{figure*}[tp!]
   \centering
   \includegraphics[width=0.8\linewidth]{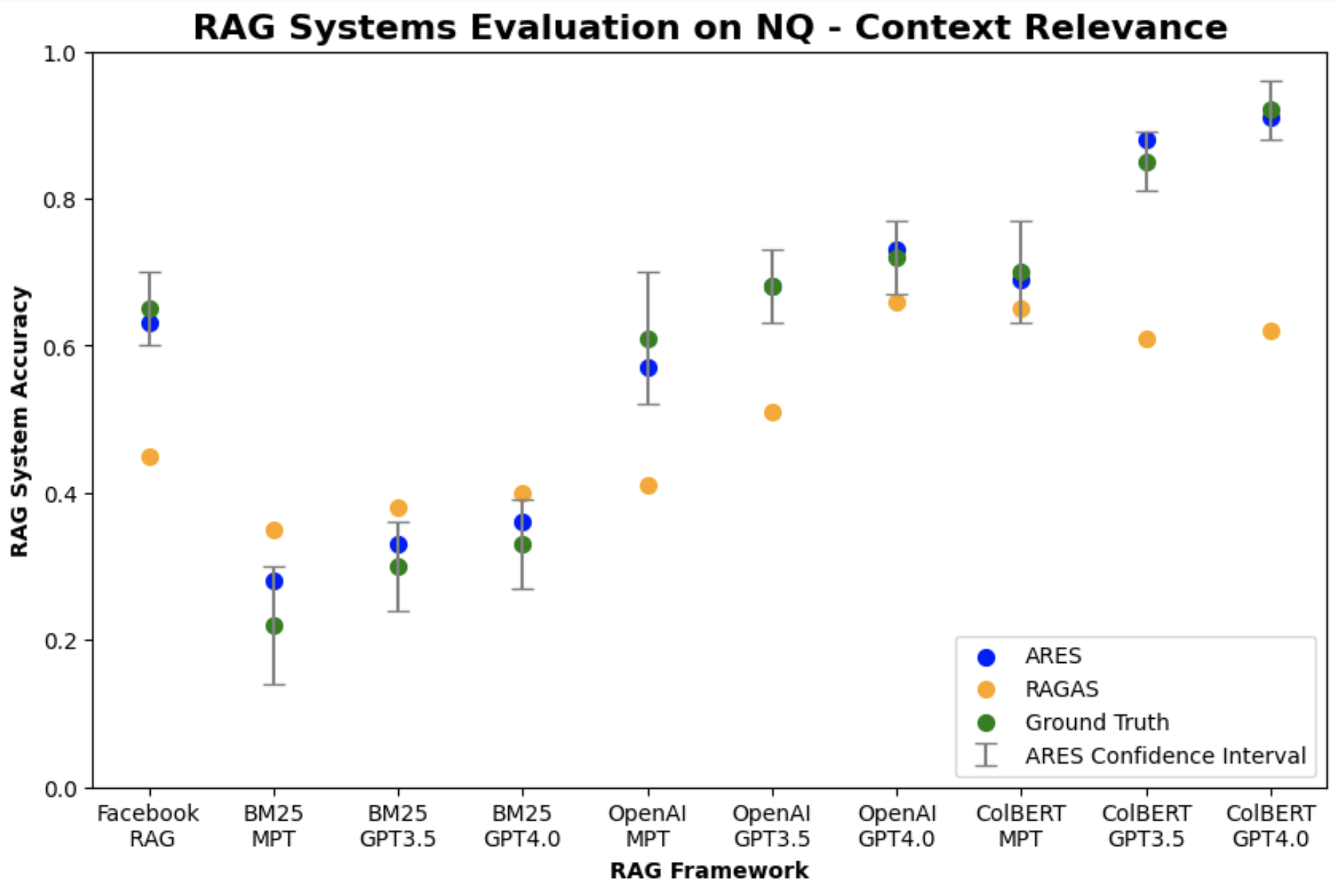}
   \caption{\textbf{RAG Systems Evaluation on NQ - Context Relevance}}
   \label{fig:nq_context_relevance_real_rag}
\end{figure*}

%% file: Tables/NQ_AnswerRelevance.tex
\begin{figure*}[tp!]
   \centering
   \includegraphics[width=0.8\linewidth]{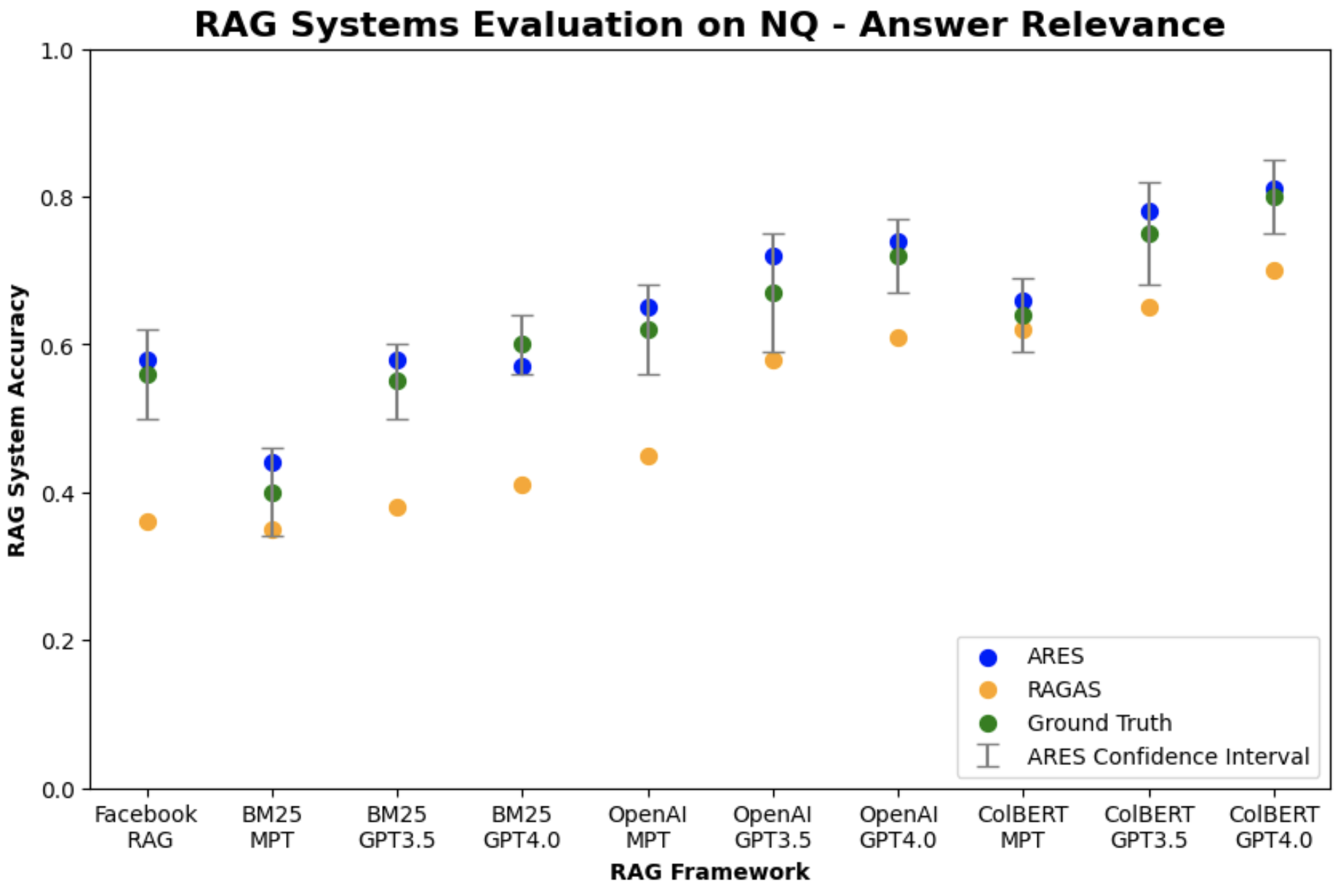}
   \caption{\textbf{RAG Systems Evaluation on NQ - Answer Relevance}}
   \label{fig:nq_answer_relevance_real_rag}
\end{figure*}

%% file: Tables/ppi_comparison_table.tex
\begin{table*}[htp!]
\small
\centering
\setlength{\tabcolsep}{3.5pt}
\begin{tabular}{crrrrrr}
\toprule
\multicolumn{1}{l}{}                                                 & \multicolumn{6}{c}{\textbf{\begin{tabular}[c]{@{}c@{}}Kendall's Tau by Dataset\end{tabular}}}                                                                                                                                                                                                                  \\
\cmidrule(l{7pt}r{7pt}){2-7}
\textbf{}                                                            & \multicolumn{2}{c}{NQ}                                                                                                                                            & \multicolumn{2}{c}{MultiRC} & \multicolumn{2}{c}{ReCoRD}                                                                                                                                        \\
\midrule
\textbf{\begin{tabular}[c]{@{}c@{}}PPI Labeled\\ Count\end{tabular}} & \multicolumn{1}{c}{\begin{tabular}[c]{@{}c@{}}C.R.\end{tabular}} & \multicolumn{1}{c}{\begin{tabular}[c]{@{}c@{}}A.R.\end{tabular}} & \multicolumn{1}{c}{\begin{tabular}[c]{@{}c@{}}C.R.\end{tabular}} & \multicolumn{1}{c}{\begin{tabular}[c]{@{}c@{}}A.R.\end{tabular}} & \multicolumn{1}{c}{\begin{tabular}[c]{@{}c@{}}C.R.\end{tabular}} & \multicolumn{1}{c}{\begin{tabular}[c]{@{}c@{}}A.R.\end{tabular}} \\
\midrule
400                                                                  & 1.0                                                                              & 1.0                                                                            & 0.89                                                                             & 0.94  & 0.89 & 0.94                                                                          \\
300                                                                  & 0.89                                                                             & 1.0                                                                            & 0.94                                                                             & 0.89   & 0.83 & 0.89                                                                         \\
200                                                                  & 0.83                                                                             & 1.0                                                                            & 0.83                                                                             & 0.94  & 0.83 &  0.83                                                                         \\
150                                                                  & 0.72                                                                             & 1.0                                                                            & 0.83                                                                             & 0.89    & 0.72 &  0.83                                                                      \\
100                                                                  & 0.44                                                                             & 1.0                                                                            & 0.67                                                                             & 0.67  & 0.67 & 0.83                                                                          \\
50                                                                   & 0.44                                                                             & 0.94                                                                           & 0.61                                                                             & 0.44   & 0.56 & 0.67                                                                         \\
25                                                                   & 0.44                                                                             & 0.89                                                                           & 0.56                                                                             & 0.44 & 0.44 & 0.56   \\ \bottomrule                                                                        
\end{tabular}
\caption{\textbf{Analysis of PPI Labeled Count vs. ARES Efficacy by Kendall's Tau}: The Kendall's tau values represent the correlation between the correct ranking and the ARES ranking of the pseudo RAG systems. We use the same experimental set-up as described in \autoref{sec:datasets}. We find that below about 100-150 datapoints in the human preference validation set, ARES cannot meaningfully distinguish between the alternate RAG systems based on their accuracies in context relevance and answer relevance (C.R. and A.R., respectively).}
\label{tab:ppi_count}
\end{table*}

%% file: Tables/GPT4_Labeling.tex
\begin{table*}[]
\small
\centering
\setlength{\tabcolsep}{2.0pt}
\begin{tabular}{lrrrrrr}
\toprule
                                                                                     & \multicolumn{6}{c}{\textbf{ARES Ranking of Pseudo RAG Systems using GPT-4 Labels}}                                                                                                                                                                                                                                                                                                                                                                                                                     \\ \cmidrule(l{7pt}r{7pt}){2-7}
                                                                                     & \multicolumn{2}{c}{NQ}                                                                                                                                           & \multicolumn{2}{c}{ReCoRD}                                                                                                                                        & \multicolumn{2}{c}{MultiRC}                                                                                                                                      \\
                                                                                     \\ \cmidrule(l{7pt}r{7pt}){2-7}
                                                                                     & \multicolumn{1}{c}{\begin{tabular}[c]{@{}c@{}}Context\\ Relevance\end{tabular}} & \multicolumn{1}{c}{\begin{tabular}[c]{@{}c@{}}Answer\\ Relevance\end{tabular}} & \multicolumn{1}{c}{\begin{tabular}[c]{@{}c@{}}Context\\ Relevance\end{tabular}} & \multicolumn{1}{c}{\begin{tabular}[c]{@{}c@{}}Answer\\ Relevance\end{tabular}} & \multicolumn{1}{c}{\begin{tabular}[c]{@{}c@{}}Context\\ Relevance\end{tabular}} & \multicolumn{1}{c}{\begin{tabular}[c]{@{}c@{}}Answer\\ Relevance\end{tabular}} 
                                                                                                                                                                                                                                                                                                                                                                                                                                                                                                                                        \\ \midrule
Kendall's Tau                                                                        & 0.78                                                                            & 1.0                                                                            & 0.78                                                                           & 0.72                                                                           & 0.89                                                                            & 0.78                                                                           \\ \midrule
\begin{tabular}[c]{@{}l@{}}Kendall's Tau of \\ Human Labeled Approach\end{tabular}                                            & 0.94                                                                            & 1.0                                                                            & 0.83                                                                            & 0.89                                                                           & 0.94                                                                            & 0.89                                                                           \\
\midrule
Average PPI Range                                                                    & 9.2\%                                                                           & 6.8\%                                                                          & 8.2\%                                                                           & 9.0\%                                                                          & 7.7\%                                                                           & 8.3\%                                                                          \\ \midrule
\begin{tabular}[c]{@{}l@{}}Accuracy on \\ RAG Evaluation Sets\end{tabular} & 79.3\%                                                                          & 96.7\%                                                                         & 88.4\%                                                                          & 78.3\%                                                                         & 85.8\%                                                                          & 82.5\%  \\ \bottomrule                                                                      
\end{tabular}
\caption{\textbf{GPT-4 Labels vs. Human Labels}: We wanted to explore the practicality of using GPT-4 generated labels instead of human annotations for our human preference validation set in ARES.
In the experiments, we generated 500 GPT-4 labels as replacements for human labeling using few-shot prompts (see Sections \ref{sec:gpt_prompting_for_context_relevance_scoring}, \ref{sec:gpt_prompting_for_answer_faithfulness_scoring}, and \ref{sec:gpt_prompting_for_answer_relevance_scoring}).
While GPT-4 generated labels decreased Kendall's tau in most settings by 0.05 to 0.30, the ability to cheaply produce GPT-4 generated labels significantly reduces the  cost of annotation, cutting it from hundreds of annotations to less than ten for few-shot prompts.
Additionally, the efficacy of PPI continues improving as we generate more GPT-4 generated labels. 
In the table, we define PPI range as the number of percentage points from the lower number to the upper number of the PPI confidence bounding. Additionally, we use the fine-tuned LLM judge (DeBERTa-v3-Large) for evaluation.}
\label{tab:gpt4_labels}
\end{table*}

%% file: Tables/ARES_on_Real_RAG.tex
\begin{table*}[]
\small
\centering
\begin{tabular}{lllllll}
\toprule
                                                                                & \multicolumn{6}{c}{\textbf{ARES Ranking of Real RAG Systems}}                                                                                                   \\
                                                                                \cmidrule(l{7pt}r{7pt}){2-7}
                                                                                & \multicolumn{2}{c}{NQ}                              & \multicolumn{2}{c}{WoW}                             & \multicolumn{2}{c}{FEVER}                           \\
                                                                                \cmidrule(l{7pt}r{7pt}){2-7}
                                                                                
\textbf{}                                                                       & \multicolumn{1}{c}{C.R.} & \multicolumn{1}{c}{A.R.} & \multicolumn{1}{c}{C.R.} & \multicolumn{1}{c}{A.R.} & \multicolumn{1}{c}{C.R.} & \multicolumn{1}{c}{A.R.} \\
\midrule
\begin{tabular}[c]{@{}l@{}}Kendall’s Tau for\\ Sampled Annotations\end{tabular} & 0.73                     & 0.78                     & 0.73                     & 0.73                     & 0.73                     & 0.82                     \\
\begin{tabular}[c]{@{}l@{}}Kendall's Tau \\ for RAGAS\end{tabular}              & 0.82                     & 0.82                     & 0.73                     & 0.82                     & 0.73                     & 0.87                     \\
\begin{tabular}[c]{@{}l@{}}Kendall's Tau\\ for GPT-3.5 Judge\end{tabular}       & 0.82                     & 0.87                     & 0.82                     & 0.82                     & 0.64                     & 0.87                     \\
\begin{tabular}[c]{@{}l@{}}Kendall's Tau\\ for ARES LLM Judge\end{tabular}      & 0.91                     & \textbf{0.96}                     & \textbf{0.91}                     & \textbf{1.0}                      & 0.73                     & 0.87                     \\
\begin{tabular}[c]{@{}l@{}}Kendall's Tau\\ for ARES\end{tabular}                & \textbf{1.0}                      & \textbf{0.96}                     & \textbf{0.91}                     & \textbf{1.0}                      & \textbf{0.82}                     & \textbf{1.0}                      \\
\midrule
RAGAS Accuracy                                                                  & 35.9\%                   & 68.2\%                   & 44.4\%                   & 80.1\%                   & 21.4\%                   & 75.9\%                   \\
GPT-3.5 Accuracy                                                                & 80.5\%                   & 91.2\%                   & 81.2\%                   & 83.5\%                   & 61.3\%                   & 54.5\%                   \\
ARES Accuracy                                                                   & 85.6\%                   & 93.3\%                   & 84.5\%                   & 88.2\%                   & 70.4\%                   & 84.0\%  \\ \bottomrule                
\end{tabular}
\caption{\textbf{ARES Ranking on Real-World RAG Systems}: 
For scoring context relevance and answer relevance (C.R. and A.R. in the table, respectively), we compare ARES with our fine-tuned LLM judges against sampled annotations benchmark, RAGAS, and a few-shot GPT-3.5 judge.
For our sampled annotations, we gather 150 annotated datapoints from each mock RAG system and use those labels to score the system.
RAGAS also uses GPT-3.5 as its judge but it uses few-shot prompts that are not targeted for each evaluation domain.
Overall, we found that ARES ranked RAG systems more accurately than RAGAS and GPT-3.5 across all the explored datasets.
Additionally, we include the Kendall's taus for the ARES LLM Judge without PPI and found that PPI further boosted the ranking accuracy of the judge across the board.
We selected GPT-3.5 instead of GPT-4 due to the lower financial costs required to run. %
For PPI in both ARES and the GPT-3.5 judge, we used 300 human annotations for our human preference validation set.
The prompts used for the GPT-3.5 judges are included in Sections \ref{sec:gpt_prompting_for_context_relevance_scoring}, \ref{sec:gpt_prompting_for_answer_faithfulness_scoring}, and \ref{sec:gpt_prompting_for_answer_relevance_scoring}.}
\label{tab:ares_real_systems}
\end{table*}

%% file: Tables/Cross_Domain.tex
\begin{table*}[]
\small
\centering
\setlength{\tabcolsep}{1.4pt}
\begin{tabular}{lrrrrrrrrrrrr}
\toprule
                                                                                     & \multicolumn{12}{c}{\textbf{ARES Cross-Domain Ranking of Pseudo RAG Systems}}                                                                                                                                                                                                                                                                                                                                                                                                                                                                                                                                                                                                                                                                                                                                                                                                                                                                                                                 \\
                                                                                     \cmidrule(l{7pt}r{7pt}){2-13}
                                                                                     & \multicolumn{2}{c}{\begin{tabular}[c]{@{}c@{}}NQ to \\ FEVER\end{tabular}}                                                                                    & \multicolumn{2}{c}{\begin{tabular}[c]{@{}c@{}}FEVER to \\ NQ\end{tabular}}                                                                                    & \multicolumn{2}{c}{\begin{tabular}[c]{@{}c@{}}NQ to \\ MultiRC\end{tabular}}                                                                                  & \multicolumn{2}{c}{\begin{tabular}[c]{@{}c@{}}MultiRC to \\ NQ\end{tabular}}                                                                                  & \multicolumn{2}{c}{\begin{tabular}[c]{@{}c@{}}NQ to \\ ReCoRD\end{tabular}}                                                                                   & \multicolumn{2}{c}{\begin{tabular}[c]{@{}c@{}}ReCoRD to \\ NQ\end{tabular}}                                                                                   \\ \cmidrule(l{7pt}r{7pt}){2-13}
                                                                                     & \multicolumn{1}{c}{C.R.}                                                      & \multicolumn{1}{c}{A.R.}                                                      & \multicolumn{1}{c}{C.R.}                                                      & \multicolumn{1}{c}{A.R.}                                                      & \multicolumn{1}{c}{C.R.}                                                      & \multicolumn{1}{c}{A.R.}                                                      & \multicolumn{1}{c}{C.R.}                                                      & \multicolumn{1}{c}{A.R.}                                                      & \multicolumn{1}{c}{C.R.}                                                      & \multicolumn{1}{c}{A.R.}                                                      & \multicolumn{1}{c}{C.R.}                                                      & \multicolumn{1}{c}{A.R.}                                                                                                                                                                                                                                                                                                                                                                                                                                                                                                                                                                                                                                                                                                                                                                                                                                                                                                                                                                                             \\
\midrule
Kendall's Tau                                                                        & 0.89                                                                          & 0.89                                                                          & 1.0                                                                           & 0.83                                                                          & 0.94                                                                          & 0.89                                                                          & 1.0                                                                           & 0.89                                                                          & 0.78                                                                          & 0.89                                                                          & 0.89                                                                          & 0.94                                                                          
\\ \midrule
\begin{tabular}[c]{@{}l@{}}Kendall's Tau of\\ In-Domain LLM Judge\end{tabular}       & 0.89                                                                          & 0.78                                                                          & 0.94                                                                          & 1.0                                                                           & 0.94                                                                          & 0.89                                                                          & 0.94                                                                          & 1.0                                                                           & 0.83                                                                          & 0.89                                                                          & 0.94                                                                          & 1.0                                                                           \\
\midrule
Average PPI Range                                                                    & 8.7\%                                                                         & 7.2\%                                                                         & 6.5\%                                                                         & 11.5\%                                                                        & 10.2\%                                                                        & 11.3\%                                                                        & 11.9\%                                                                        & 11.5\%                                                                        & 10.5\%                                                                        & 10.1\%                                                                        & 9.7\%                                                                         & 6.2\%                                                                         \\
\midrule
\begin{tabular}[c]{@{}l@{}} Accuracy on \\ RAG Evaluation Sets\end{tabular} & 92.4\%                                                                        & 28.4\%                                                                        & 85.7\%                                                                        & 22.6\%                                                                        & 81.5\%                                                                        & 92.1\%                                                                        & 87.6\%                                                                        & 80.2\%                                                                        & 29.1\%                                                                        & 81.2\%                                                                        & 80.1\%                                                                        & 92.1\%                                                                       \\ \bottomrule
\end{tabular}
\caption{\textbf{Cross-Domain Usage of Fine-tuned LLM Judges}: We tested the cross-domain application of the fine-tuned LLM judge in the ARES framework. We found that for both context relevance and answer relevance (C.R. and A.R. in the table, respectively), fine-tuned LLM judges showed strong generalizability across domains when changing query type (e.g. NQ and FEVER), document type (e.g. NQ and MultiRC), or both (e.g. NQ and ReCoRD). For PPI, we used 300 labeled examples for our human preference validation set but also found that additional examples further improved the performance of ARES. Furthermore, we found that even in scenarios where the fine-tuned LLM judge's accuracy significantly dropped out-of-domain (e.g. answer relevance for NQ to FEVER), PPI mitigated the decrease in judge performance. In the table, we define PPI range as the number of percentage points from the lower bound to the upper bound of the PPI confidence interval.}
\label{tab:cross_domain}
\end{table*}

%% file: acl_latex.bbl
\begin{thebibliography}{47}
\expandafter\ifx\csname natexlab\endcsname\relax\def\natexlab#1{#1}\fi

\bibitem[{Akhtar et~al.(2023)Akhtar, Aly, Christodoulopoulos, Cocarascu, Guo, Mittal, Schlichtkrull, Thorne, and Vlachos}]{fever-2023-fact}
Mubashara Akhtar, Rami Aly, Christos Christodoulopoulos, Oana Cocarascu, Zhijiang Guo, Arpit Mittal, Michael Schlichtkrull, James Thorne, and Andreas Vlachos, editors. 2023.
\newblock \href {https://aclanthology.org/2023.fever-1.0} {\emph{Proceedings of the Sixth Fact Extraction and VERification Workshop (FEVER)}}. Association for Computational Linguistics, Dubrovnik, Croatia.

\bibitem[{Angelopoulos et~al.(2023)Angelopoulos, Bates, Fannjiang, Jordan, and Zrnic}]{angelopoulos2023predictionpowered}
Anastasios~N. Angelopoulos, Stephen Bates, Clara Fannjiang, Michael~I. Jordan, and Tijana Zrnic. 2023.
\newblock \href {http://arxiv.org/abs/2301.09633} {Prediction-powered inference}.

\bibitem[{Brown et~al.(2020)Brown, Mann, Ryder, Subbiah, Kaplan, Dhariwal, Neelakantan, Shyam, Sastry, Askell, Agarwal, Herbert-Voss, Krueger, Henighan, Child, Ramesh, Ziegler, Wu, Winter, Hesse, Chen, Sigler, Litwin, Gray, Chess, Clark, Berner, McCandlish, Radford, Sutskever, and Amodei}]{brown2020language}
Tom~B. Brown, Benjamin Mann, Nick Ryder, Melanie Subbiah, Jared Kaplan, Prafulla Dhariwal, Arvind Neelakantan, Pranav Shyam, Girish Sastry, Amanda Askell, Sandhini Agarwal, Ariel Herbert-Voss, Gretchen Krueger, Tom Henighan, Rewon Child, Aditya Ramesh, Daniel~M. Ziegler, Jeffrey Wu, Clemens Winter, Christopher Hesse, Mark Chen, Eric Sigler, Mateusz Litwin, Scott Gray, Benjamin Chess, Jack Clark, Christopher Berner, Sam McCandlish, Alec Radford, Ilya Sutskever, and Dario Amodei. 2020.
\newblock \href {http://arxiv.org/abs/2005.14165} {Language models are few-shot learners}.

\bibitem[{Chen et~al.(2023)Chen, Lin, Han, and Sun}]{chen2023benchmarking}
Jiawei Chen, Hongyu Lin, Xianpei Han, and Le~Sun. 2023.
\newblock Benchmarking large language models in retrieval-augmented generation.
\newblock \emph{arXiv preprint arXiv:2309.01431}.

\bibitem[{Chung et~al.(2022)Chung, Hou, Longpre, Zoph, Tay, Fedus, Li, Wang, Dehghani, Brahma et~al.}]{chung2022scaling}
Hyung~Won Chung, Le~Hou, Shayne Longpre, Barret Zoph, Yi~Tay, William Fedus, Eric Li, Xuezhi Wang, Mostafa Dehghani, Siddhartha Brahma, et~al. 2022.
\newblock Scaling instruction-finetuned language models.
\newblock \emph{arXiv preprint arXiv:2210.11416}.

\bibitem[{Dai et~al.(2022)Dai, Zhao, Ma, Luan, Ni, Lu, Bakalov, Guu, Hall, and Chang}]{dai2022promptagator}
Zhuyun Dai, Vincent~Y Zhao, Ji~Ma, Yi~Luan, Jianmo Ni, Jing Lu, Anton Bakalov, Kelvin Guu, Keith~B Hall, and Ming-Wei Chang. 2022.
\newblock Promptagator: Few-shot dense retrieval from 8 examples.
\newblock \emph{arXiv preprint arXiv:2209.11755}.

\bibitem[{Dinan et~al.(2018)Dinan, Roller, Shuster, Fan, Auli, and Weston}]{dinan2018wizard}
Emily Dinan, Stephen Roller, Kurt Shuster, Angela Fan, Michael Auli, and Jason Weston. 2018.
\newblock Wizard of wikipedia: Knowledge-powered conversational agents.
\newblock \emph{arXiv preprint arXiv:1811.01241}.

\bibitem[{Elsahar et~al.(2018)Elsahar, Vougiouklis, Remaci, Gravier, Hare, Laforest, and Simperl}]{elsahar2018t}
Hady Elsahar, Pavlos Vougiouklis, Arslen Remaci, Christophe Gravier, Jonathon Hare, Frederique Laforest, and Elena Simperl. 2018.
\newblock T-rex: A large scale alignment of natural language with knowledge base triples.
\newblock In \emph{Proceedings of the Eleventh International Conference on Language Resources and Evaluation (LREC 2018)}.

\bibitem[{Fu et~al.(2023)Fu, Ng, Jiang, and Liu}]{fu2023gptscore}
Jinlan Fu, See-Kiong Ng, Zhengbao Jiang, and Pengfei Liu. 2023.
\newblock Gptscore: Evaluate as you desire.
\newblock \emph{arXiv preprint arXiv:2302.04166}.

\bibitem[{Gao et~al.(2023)Gao, Yen, Yu, and Chen}]{gao2023enabling}
Tianyu Gao, Howard Yen, Jiatong Yu, and Danqi Chen. 2023.
\newblock \href {http://arxiv.org/abs/2305.14627} {Enabling large language models to generate text with citations}.

\bibitem[{Gekhman et~al.(2023)Gekhman, Herzig, Aharoni, Elkind, and Szpektor}]{gekhman2023trueteacher}
Zorik Gekhman, Jonathan Herzig, Roee Aharoni, Chen Elkind, and Idan Szpektor. 2023.
\newblock \href {http://arxiv.org/abs/2305.11171} {Trueteacher: Learning factual consistency evaluation with large language models}.

\bibitem[{Guu et~al.(2020)Guu, Lee, Tung, Pasupat, and Chang}]{guu2020retrieval}
Kelvin Guu, Kenton Lee, Zora Tung, Panupong Pasupat, and Mingwei Chang. 2020.
\newblock Retrieval augmented language model pre-training.
\newblock In \emph{International conference on machine learning}, pages 3929--3938. PMLR.

\bibitem[{He et~al.(2021)He, Gao, and Chen}]{he2021debertav3}
Pengcheng He, Jianfeng Gao, and Weizhu Chen. 2021.
\newblock Debertav3: Improving deberta using electra-style pre-training with gradient-disentangled embedding sharing.
\newblock \emph{arXiv preprint arXiv:2111.09543}.

\bibitem[{Howard and Ruder(2018)}]{howard-ruder-2018-universal}
Jeremy Howard and Sebastian Ruder. 2018.
\newblock \href {https://doi.org/10.18653/v1/P18-1031} {Universal language model fine-tuning for text classification}.
\newblock In \emph{Proceedings of the 56th Annual Meeting of the Association for Computational Linguistics (Volume 1: Long Papers)}, pages 328--339, Melbourne, Australia. Association for Computational Linguistics.

\bibitem[{Huo et~al.(2023)Huo, Arabzadeh, and Clarke}]{huo2023retrieving}
Siqing Huo, Negar Arabzadeh, and Charles~LA Clarke. 2023.
\newblock Retrieving supporting evidence for llms generated answers.
\newblock \emph{arXiv preprint arXiv:2306.13781}.

\bibitem[{Husain et~al.(2019)Husain, Wu, Gazit, Allamanis, and Brockschmidt}]{husain2019codesearchnet}
Hamel Husain, Ho-Hsiang Wu, Tiferet Gazit, Miltiadis Allamanis, and Marc Brockschmidt. 2019.
\newblock {CodeSearchNet} challenge: Evaluating the state of semantic code search.
\newblock \emph{arXiv preprint arXiv:1909.09436}.

\bibitem[{Izacard et~al.(2022)Izacard, Lewis, Lomeli, Hosseini, Petroni, Schick, Dwivedi-Yu, Joulin, Riedel, and Grave}]{izacard2022few}
Gautier Izacard, Patrick Lewis, Maria Lomeli, Lucas Hosseini, Fabio Petroni, Timo Schick, Jane Dwivedi-Yu, Armand Joulin, Sebastian Riedel, and Edouard Grave. 2022.
\newblock Few-shot learning with retrieval augmented language models.
\newblock \emph{arXiv preprint arXiv:2208.03299}.

\bibitem[{James and Es(2023)}]{ragas2023}
Jithin James and Shahul Es. 2023.
\newblock \href {https://github.com/explodinggradients/ragas} {Ragas: Evaluation framework for your retrieval augmented generation (rag) pipelines}.

\bibitem[{Johnson et~al.(2019)Johnson, Douze, and J{\'e}gou}]{johnson2019billion}
Jeff Johnson, Matthijs Douze, and Herv{\'e} J{\'e}gou. 2019.
\newblock Billion-scale similarity search with {GPUs}.
\newblock \emph{IEEE Transactions on Big Data}, 7(3):535--547.

\bibitem[{Kamalloo et~al.(2023)Kamalloo, Jafari, Zhang, Thakur, and Lin}]{kamalloo2023hagrid}
Ehsan Kamalloo, Aref Jafari, Xinyu Zhang, Nandan Thakur, and Jimmy Lin. 2023.
\newblock \href {http://arxiv.org/abs/2307.16883} {Hagrid: A human-llm collaborative dataset for generative information-seeking with attribution}.

\bibitem[{Karpukhin et~al.(2020)Karpukhin, Oğuz, Min, Lewis, Wu, Edunov, Chen, and tau Yih}]{karpukhin2020dense}
Vladimir Karpukhin, Barlas Oğuz, Sewon Min, Patrick Lewis, Ledell Wu, Sergey Edunov, Danqi Chen, and Wen tau Yih. 2020.
\newblock \href {http://arxiv.org/abs/2004.04906} {Dense passage retrieval for open-domain question answering}.

\bibitem[{Khashabi et~al.(2018)Khashabi, Chaturvedi, Roth, Upadhyay, and Roth}]{khashabi2018looking}
Daniel Khashabi, Snigdha Chaturvedi, Michael Roth, Shyam Upadhyay, and Dan Roth. 2018.
\newblock Looking beyond the surface: A challenge set for reading comprehension over multiple sentences.
\newblock In \emph{Proceedings of the 2018 Conference of the North American Chapter of the Association for Computational Linguistics: Human Language Technologies, Volume 1 (Long Papers)}, pages 252--262.

\bibitem[{Khattab et~al.(2021)Khattab, Potts, and Zaharia}]{khattab2021relevance}
Omar Khattab, Christopher Potts, and Matei Zaharia. 2021.
\newblock Relevance-guided supervision for openqa with colbert.
\newblock \emph{Transactions of the association for computational linguistics}, 9:929--944.

\bibitem[{Kingma and Ba(2017)}]{kingma2017adam}
Diederik~P. Kingma and Jimmy Ba. 2017.
\newblock \href {http://arxiv.org/abs/1412.6980} {Adam: A method for stochastic optimization}.

\bibitem[{Kocmi and Federmann(2023)}]{kocmi2023large}
Tom Kocmi and Christian Federmann. 2023.
\newblock Large language models are state-of-the-art evaluators of translation quality.
\newblock \emph{arXiv preprint arXiv:2302.14520}.

\bibitem[{Krishna et~al.(2023)Krishna, Bransom, Kuehl, Iyyer, Dasigi, Cohan, and Lo}]{krishna-etal-2023-longeval}
Kalpesh Krishna, Erin Bransom, Bailey Kuehl, Mohit Iyyer, Pradeep Dasigi, Arman Cohan, and Kyle Lo. 2023.
\newblock \href {https://aclanthology.org/2023.eacl-main.121} {{L}ong{E}val: Guidelines for human evaluation of faithfulness in long-form summarization}.
\newblock In \emph{Proceedings of the 17th Conference of the European Chapter of the Association for Computational Linguistics}, pages 1650--1669, Dubrovnik, Croatia. Association for Computational Linguistics.

\bibitem[{Kwiatkowski et~al.(2019)Kwiatkowski, Palomaki, Redfield, Collins, Parikh, Alberti, Epstein, Polosukhin, Devlin, Lee et~al.}]{kwiatkowski2019natural}
Tom Kwiatkowski, Jennimaria Palomaki, Olivia Redfield, Michael Collins, Ankur Parikh, Chris Alberti, Danielle Epstein, Illia Polosukhin, Jacob Devlin, Kenton Lee, et~al. 2019.
\newblock Natural questions: a benchmark for question answering research.
\newblock \emph{Transactions of the Association for Computational Linguistics}, 7:453--466.

\bibitem[{Lewis et~al.(2019)Lewis, Liu, Goyal, Ghazvininejad, Mohamed, Levy, Stoyanov, and Zettlemoyer}]{lewis2019bart}
Mike Lewis, Yinhan Liu, Naman Goyal, Marjan Ghazvininejad, Abdelrahman Mohamed, Omer Levy, Ves Stoyanov, and Luke Zettlemoyer. 2019.
\newblock \href {http://arxiv.org/abs/1910.13461} {Bart: Denoising sequence-to-sequence pre-training for natural language generation, translation, and comprehension}.

\bibitem[{Lewis et~al.(2020)Lewis, Perez, Piktus, Petroni, Karpukhin, Goyal, K{\"u}ttler, Lewis, Yih, Rockt{\"a}schel et~al.}]{lewis2020retrieval}
Patrick Lewis, Ethan Perez, Aleksandra Piktus, Fabio Petroni, Vladimir Karpukhin, Naman Goyal, Heinrich K{\"u}ttler, Mike Lewis, Wen-tau Yih, Tim Rockt{\"a}schel, et~al. 2020.
\newblock Retrieval-augmented generation for knowledge-intensive nlp tasks.
\newblock \emph{Advances in Neural Information Processing Systems}, 33:9459--9474.

\bibitem[{Liang et~al.(2020)Liang, Duan, Gong, Wu, Guo, Qi, Gong, Shou, Jiang, Cao, Fan, Zhang, Agrawal, Cui, Wei, Bharti, Qiao, Chen, Wu, Liu, Yang, Campos, Majumder, and Zhou}]{Liang2020XGLUEAN}
Yaobo Liang, Nan Duan, Yeyun Gong, Ning Wu, Fenfei Guo, Weizhen Qi, Ming Gong, Linjun Shou, Daxin Jiang, Guihong Cao, Xiaodong Fan, Ruofei Zhang, Rahul Agrawal, Edward Cui, Sining Wei, Taroon Bharti, Ying Qiao, Jiun-Hung Chen, Winnie Wu, Shuguang Liu, Fan Yang, Daniel Campos, Rangan Majumder, and Ming Zhou. 2020.
\newblock Xglue: A new benchmark dataset for cross-lingual pre-training, understanding and generation.
\newblock \emph{arXiv}, abs/2004.01401.

\bibitem[{Liu et~al.(2023{\natexlab{a}})Liu, Iter, Xu, Wang, Xu, and Zhu}]{liu2303g}
Yang Liu, Dan Iter, Yichong Xu, Shuohang Wang, Ruochen Xu, and Chenguang Zhu. 2023{\natexlab{a}}.
\newblock G-eval: Nlg evaluation using gpt-4 with better human alignment, may 2023.
\newblock \emph{arXiv preprint arXiv:2303.16634}.

\bibitem[{Liu et~al.(2023{\natexlab{b}})Liu, Yang, Huang, Zhang, Huang, Wei, Deng, Sun, and Zhang}]{liu2023calibrating}
Yuxuan Liu, Tianchi Yang, Shaohan Huang, Zihan Zhang, Haizhen Huang, Furu Wei, Weiwei Deng, Feng Sun, and Qi~Zhang. 2023{\natexlab{b}}.
\newblock Calibrating llm-based evaluator.
\newblock \emph{arXiv preprint arXiv:2309.13308}.

\bibitem[{Mialon et~al.(2023)Mialon, Dessì, Lomeli, Nalmpantis, Pasunuru, Raileanu, Rozière, Schick, Dwivedi-Yu, Celikyilmaz, Grave, LeCun, and Scialom}]{mialon2023augmented}
Grégoire Mialon, Roberto Dessì, Maria Lomeli, Christoforos Nalmpantis, Ram Pasunuru, Roberta Raileanu, Baptiste Rozière, Timo Schick, Jane Dwivedi-Yu, Asli Celikyilmaz, Edouard Grave, Yann LeCun, and Thomas Scialom. 2023.
\newblock \href {http://arxiv.org/abs/2302.07842} {Augmented language models: a survey}.

\bibitem[{Min et~al.(2023)Min, Krishna, Lyu, Lewis, tau Yih, Koh, Iyyer, Zettlemoyer, and Hajishirzi}]{min2023factscore}
Sewon Min, Kalpesh Krishna, Xinxi Lyu, Mike Lewis, Wen tau Yih, Pang~Wei Koh, Mohit Iyyer, Luke Zettlemoyer, and Hannaneh Hajishirzi. 2023.
\newblock \href {http://arxiv.org/abs/2305.14251} {Factscore: Fine-grained atomic evaluation of factual precision in long form text generation}.

\bibitem[{Petroni et~al.(2021)Petroni, Piktus, Fan, Lewis, Yazdani, De~Cao, Thorne, Jernite, Karpukhin, Maillard, Plachouras, Rockt{\"a}schel, and Riedel}]{petroni-etal-2021-kilt}
Fabio Petroni, Aleksandra Piktus, Angela Fan, Patrick Lewis, Majid Yazdani, Nicola De~Cao, James Thorne, Yacine Jernite, Vladimir Karpukhin, Jean Maillard, Vassilis Plachouras, Tim Rockt{\"a}schel, and Sebastian Riedel. 2021.
\newblock \href {https://doi.org/10.18653/v1/2021.naacl-main.200} {{KILT}: a benchmark for knowledge intensive language tasks}.
\newblock In \emph{Proceedings of the 2021 Conference of the North American Chapter of the Association for Computational Linguistics: Human Language Technologies}, pages 2523--2544, Online. Association for Computational Linguistics.

\bibitem[{Rashkin et~al.(2022)Rashkin, Nikolaev, Lamm, Aroyo, Collins, Das, Petrov, Tomar, Turc, and Reitter}]{rashkin2022measuring}
Hannah Rashkin, Vitaly Nikolaev, Matthew Lamm, Lora Aroyo, Michael Collins, Dipanjan Das, Slav Petrov, Gaurav~Singh Tomar, Iulia Turc, and David Reitter. 2022.
\newblock \href {http://arxiv.org/abs/2112.12870} {Measuring attribution in natural language generation models}.

\bibitem[{Saad-Falcon et~al.(2023)Saad-Falcon, Khattab, Santhanam, Florian, Franz, Roukos, Sil, Sultan, and Potts}]{saad2023udapdr}
Jon Saad-Falcon, Omar Khattab, Keshav Santhanam, Radu Florian, Martin Franz, Salim Roukos, Avirup Sil, Md~Arafat Sultan, and Christopher Potts. 2023.
\newblock Udapdr: Unsupervised domain adaptation via llm prompting and distillation of rerankers.
\newblock \emph{arXiv preprint arXiv:2303.00807}.

\bibitem[{Sander and Dietz(2021)}]{sander2021exam}
David~P Sander and Laura Dietz. 2021.
\newblock Exam: How to evaluate retrieve-and-generate systems for users who do not (yet) know what they want.
\newblock In \emph{DESIRES}, pages 136--146.

\bibitem[{Santhanam et~al.(2022)Santhanam, Khattab, Saad-Falcon, Potts, and Zaharia}]{santhanam-etal-2022-colbertv2}
Keshav Santhanam, Omar Khattab, Jon Saad-Falcon, Christopher Potts, and Matei Zaharia. 2022.
\newblock \href {https://doi.org/10.18653/v1/2022.naacl-main.272} {{C}ol{BERT}v2: Effective and efficient retrieval via lightweight late interaction}.
\newblock In \emph{Proceedings of the 2022 Conference of the North American Chapter of the Association for Computational Linguistics: Human Language Technologies}, pages 3715--3734, Seattle, United States. Association for Computational Linguistics.

\bibitem[{Shuster et~al.(2021)Shuster, Poff, Chen, Kiela, and Weston}]{shuster2021retrieval}
Kurt Shuster, Spencer Poff, Moya Chen, Douwe Kiela, and Jason Weston. 2021.
\newblock \href {http://arxiv.org/abs/2104.07567} {Retrieval augmentation reduces hallucination in conversation}.

\bibitem[{Team(2023)}]{MosaicML2023Introducing}
MosaicML~NLP Team. 2023.
\newblock \href {www.mosaicml.com/blog/mpt-30b} {Introducing mpt-30b: Raising the bar for open-source foundation models}.
\newblock Accessed: 2023-06-22.

\bibitem[{Wang et~al.(2019)Wang, Pruksachatkun, Nangia, Singh, Michael, Hill, Levy, and Bowman}]{wang2019superglue}
Alex Wang, Yada Pruksachatkun, Nikita Nangia, Amanpreet Singh, Julian Michael, Felix Hill, Omer Levy, and Samuel Bowman. 2019.
\newblock Superglue: A stickier benchmark for general-purpose language understanding systems.
\newblock \emph{Advances in neural information processing systems}, 32.

\bibitem[{Wang et~al.(2023)Wang, Liang, Meng, Shi, Li, Xu, Qu, and Zhou}]{wang2023chatgpt}
Jiaan Wang, Yunlong Liang, Fandong Meng, Haoxiang Shi, Zhixu Li, Jinan Xu, Jianfeng Qu, and Jie Zhou. 2023.
\newblock Is chatgpt a good nlg evaluator? a preliminary study.
\newblock \emph{arXiv preprint arXiv:2303.04048}.

\bibitem[{Yang et~al.(2018)Yang, Qi, Zhang, Bengio, Cohen, Salakhutdinov, and Manning}]{yang2018hotpotqa}
Zhilin Yang, Peng Qi, Saizheng Zhang, Yoshua Bengio, William~W Cohen, Ruslan Salakhutdinov, and Christopher~D Manning. 2018.
\newblock Hotpotqa: A dataset for diverse, explainable multi-hop question answering.
\newblock \emph{arXiv preprint arXiv:1809.09600}.

\bibitem[{Yue et~al.(2023)Yue, Wang, Chen, Zhang, Su, and Sun}]{yue2023automatic}
Xiang Yue, Boshi Wang, Ziru Chen, Kai Zhang, Yu~Su, and Huan Sun. 2023.
\newblock \href {http://arxiv.org/abs/2305.06311} {Automatic evaluation of attribution by large language models}.

\bibitem[{Zhang et~al.(2018)Zhang, Liu, Liu, Gao, Duh, and Van~Durme}]{zhang2018record}
Sheng Zhang, Xiaodong Liu, Jingjing Liu, Jianfeng Gao, Kevin Duh, and Benjamin Van~Durme. 2018.
\newblock Record: Bridging the gap between human and machine commonsense reading comprehension.
\newblock \emph{arXiv preprint arXiv:1810.12885}.

\bibitem[{Zheng et~al.(2023)Zheng, Chiang, Sheng, Zhuang, Wu, Zhuang, Lin, Li, Li, Xing et~al.}]{zheng2023judging}
Lianmin Zheng, Wei-Lin Chiang, Ying Sheng, Siyuan Zhuang, Zhanghao Wu, Yonghao Zhuang, Zi~Lin, Zhuohan Li, Dacheng Li, Eric Xing, et~al. 2023.
\newblock Judging llm-as-a-judge with mt-bench and chatbot arena.
\newblock \emph{arXiv preprint arXiv:2306.05685}.

\end{thebibliography}
